\documentclass[a4paper,11pt]{article}
\usepackage[T1]{fontenc} 
\usepackage{float} 
\usepackage{booktabs}
\usepackage{multirow}
\usepackage{diagbox}

\makeatletter
\gdef\@fpheader{ }
\gdef\@journal{ }
\makeatother
\RequirePackage{amsmath}
\usepackage{txfonts}
\RequirePackage{amssymb}
\RequirePackage{epsfig}
\RequirePackage{graphicx}
\RequirePackage[numbers,sort&compress]{natbib}
\RequirePackage{color}
\RequirePackage[colorlinks=true
,urlcolor=blue
,anchorcolor=blue
,citecolor=blue
,filecolor=blue
,linkcolor=blue
,menucolor=blue
,pagecolor=blue
,linktocpage=true
,pdfproducer=medialab
,pdfa=true
]{hyperref}

\newif\ifnotoc\notocfalse
\newif\ifemailadd\emailaddfalse
\newif\iftoccontinuous\toccontinuousfalse
\makeatletter
\def\@subheader{\@empty}
\def\@keywords{\@empty}
\def\@abstract{\@empty}
\def\@xtum{\@empty}
\def\@dedicated{\@empty}
\def\@arxivnumber{\@empty}
\def\@collaboration{\@empty}
\def\@collaborationImg{\@empty}
\def\@proceeding{\@empty}
\def\@preprint{\@empty}

\newcommand{\subheader}[1]{\gdef\@subheader{#1}}
\newcommand{\keywords}[1]{\if!\@keywords!\gdef\@keywords{#1}\else%
\PackageWarningNoLine{\jname}{Keywords already defined.\MessageBreak Ignoring last definition.}\fi}
\renewcommand{\abstract}[1]{\gdef\@abstract{#1}}
\newcommand{\dedicated}[1]{\gdef\@dedicated{#1}}
\newcommand{\arxivnumber}[1]{\gdef\@arxivnumber{#1}}
\newcommand{\proceeding}[1]{\gdef\@proceeding{#1}}
\newcommand{\xtumfont}[1]{\textsc{#1}}
\newcommand{\correctionref}[3]{\gdef\@xtum{\xtumfont{#1} \href{#2}{#3}}}
\newcommand\jname{JHEP}

\newcommand\preprint[1]{\gdef\@preprint{\hfill #1}}

\makeatother

\newcommand\note[2][]{%
\if!#1!%
\stepcounter{footnote}\footnotetext{#2}%
\else%
{\renewcommand\thefootnote{#1}%
\footnotetext{#2}}%
\fi}

\makeatletter
\newtoks\auth@toks
\renewcommand{\author}[2][]{%
  \if!#1!%
    \auth@toks=\expandafter{\the\auth@toks#2\ }%
  \else
    \auth@toks=\expandafter{\the\auth@toks#2$^{#1}$\ }%
  \fi
}
\makeatother
\makeatletter
\newtoks\affil@toks\newif\ifaffil\affilfalse
\newcommand{\affiliation}[2][]{%
\affiltrue
  \if!#1!%
    \affil@toks=\expandafter{\the\affil@toks{\item[]#2}}%
  \else
    \affil@toks=\expandafter{\the\affil@toks{\item[$^{#1}$]#2}}%
  \fi
}
\makeatother
\makeatletter
\newtoks\email@toks\newcounter{email@counter}%
\newcommand{\emailAdd}[1]{%
\emailaddtrue%
\ifnum\theemail@counter>0\email@toks=\expandafter{\the\email@toks, \@email{#1}}%
\else\email@toks=\expandafter{\the\email@toks\@email{#1}}%
\fi\stepcounter{email@counter}}
\newcommand{\@email}[1]{\href{mailto:#1}{\tt #1}}
\makeatother

\makeatletter
\newcommand*\collaboration[1]{\gdef\@collaboration{#1}}
\newcommand*\collaborationImg[2][]{\gdef\@collaborationImg{#2}}
\makeatletter
\newcommand\afterLogoSpace{\smallskip}
\newcommand\afterSubheaderSpace{\vskip3pt plus 2pt minus 1pt}
\newcommand\afterProceedingsSpace{\vskip21pt plus0.4fil minus15pt}
\newcommand\afterTitleSpace{\vskip23pt plus0.06fil minus13pt}
\newcommand\afterRuleSpace{\vskip23pt plus0.06fil minus13pt}
\newcommand\afterCollaborationSpace{\vskip3pt plus 2pt minus 1pt}
\newcommand\afterCollaborationImgSpace{\vskip3pt plus 2pt minus 1pt}
\newcommand\afterAuthorSpace{\vskip5pt plus4pt minus4pt}
\newcommand\afterAffiliationSpace{\vskip3pt plus3pt}
\newcommand\afterEmailSpace{\vskip16pt plus9pt minus10pt\filbreak}
\newcommand\afterXtumSpace{\par\bigskip}
\newcommand\afterAbstractSpace{\vskip16pt plus9pt minus13pt}
\newcommand\afterKeywordsSpace{\vskip16pt plus9pt minus13pt}
\newcommand\afterArxivSpace{\vskip3pt plus0.01fil minus10pt}
\newcommand\afterDedicatedSpace{\vskip0pt plus0.01fil}
\newcommand\afterTocSpace{\bigskip\medskip}
\newcommand\afterTocRuleSpace{\bigskip\bigskip}
\newlength{\affiliationsSep}
\newcommand\beforetochook{\pagestyle{myplain}\pagenumbering{roman}}

\DeclareFixedFont\trfont{OT1}{phv}{b}{sc}{11}

\renewcommand\maketitle{
\pagestyle{empty}
\thispagestyle{titlepage}
\setcounter{page}{0}
\noindent{\small\scshape\@fpheader}\@preprint\par

\afterLogoSpace
\if!\@subheader!\else\noindent{\trfont{\@subheader}}\fi
\afterSubheaderSpace
\if!\@proceeding!\else\noindent{\sc\@proceeding}\fi
\afterProceedingsSpace
{\LARGE\flushleft\sffamily\bfseries\@title\par}
\afterTitleSpace
\hrule height 1.5\p@%
\afterRuleSpace
\if!\@collaboration!\else
{\Large\bfseries\sffamily\raggedright\@collaboration}\par
\afterCollaborationSpace
\fi
\if!\@collaborationImg!\else
{\normalsize\bfseries\sffamily\raggedright\@collaborationImg}\par
\afterCollaborationImgSpace
\fi
{\bfseries\raggedright\sffamily\the\auth@toks\par}
\afterAuthorSpace
\ifaffil\begin{list}{}{%
\setlength{\leftmargin}{0.28cm}%
\setlength{\labelsep}{0pt}%
\setlength{\itemsep}{\affiliationsSep}%
\setlength{\topsep}{-\parskip}}
\itshape\small%
\the\affil@toks
\end{list}\fi
\afterAffiliationSpace
\ifemailadd 
\noindent\hspace{0.28cm}\begin{minipage}[l]{.9\textwidth}
\begin{flushleft}
\textit{E-mail:} \the\email@toks
\end{flushleft}
\end{minipage}
\else 
\PackageWarningNoLine{\jname}{E-mails are missing.\MessageBreak Plese use \protect\emailAdd\space macro to provide e-mails.}
\fi
\afterEmailSpace
\if!\@xtum!\else\noindent{\@xtum}\afterXtumSpace\fi
\if!\@abstract!\else\noindent{\renewcommand\baselinestretch{.9}\textsc{Abstract:}}\ \@abstract\afterAbstractSpace\fi
\if!\@keywords!\else\noindent{\textsc{Keywords:}} \@keywords\afterKeywordsSpace\fi
\if!\@arxivnumber!\else\noindent{\textsc{ArXiv ePrint:}} \href{http://arxiv.org/abs/\@arxivnumber}{\@arxivnumber}\afterArxivSpace\fi
\if!\@dedicated!\else\vbox{\small\it\raggedleft\@dedicated}\afterDedicatedSpace\fi
\ifnotoc\else
\iftoccontinuous\else\newpage\fi
\beforetochook\hrule
\tableofcontents
\afterTocSpace
\hrule
\afterTocRuleSpace
\fi
\setcounter{footnote}{0}
\pagestyle{myplain}\pagenumbering{arabic}
} 

\renewcommand{\baselinestretch}{1.1}\normalsize
\divide\@tempdima\baselineskip
\@tempcnta=\@tempdima
\skip\@mpfootins = \skip\footins
\renewcommand{\@dotsep}{10000}

\newcommand\ps@myplain{
\pagenumbering{arabic}
\renewcommand\@oddfoot{\hfill-- \thepage\ --\hfill}
\renewcommand\@oddhead{}}
\let\ps@plain=\ps@myplain

\newcommand\ps@titlepage{\renewcommand\@oddfoot{}\renewcommand\@oddhead{}}

\numberwithin{equation}{section}

\renewcommand\section{\@startsection{section}{1}{\z@}%
                                   {-3.5ex \@plus -1.3ex \@minus -.7ex}%
                                   {2.3ex \@plus.4ex \@minus .4ex}%
                                   {\normalfont\large\bfseries}}
\renewcommand\subsection{\@startsection{subsection}{2}{\z@}%
                                   {-2.3ex\@plus -1ex \@minus -.5ex}%
                                   {1.2ex \@plus .3ex \@minus .3ex}%
                                   {\normalfont\normalsize\bfseries}}
\renewcommand\subsubsection{\@startsection{subsubsection}{3}{\z@}%
                                   {-2.3ex\@plus -1ex \@minus -.5ex}%
                                   {1ex \@plus .2ex \@minus .2ex}%
                                   {\normalfont\normalsize\bfseries}}
\renewcommand\paragraph{\@startsection{paragraph}{4}{\z@}%
                                   {1.75ex \@plus1ex \@minus.2ex}%
                                   {-1em}%
                                   {\normalfont\normalsize\bfseries}}
\renewcommand\subparagraph{\@startsection{subparagraph}{5}{\parindent}%
                                   {1.75ex \@plus1ex \@minus .2ex}%
                                   {-1em}%
                                   {\normalfont\normalsize\bfseries}}

\def\fnum@figure{\textbf{\figurename\nobreakspace\thefigure}}
\def\fnum@table{\textbf{\tablename\nobreakspace\thetable}}

\long\def\@makecaption#1#2{%
  \vskip\abovecaptionskip
  \sbox\@tempboxa{\small #1. #2}%
  \ifdim \wd\@tempboxa >\hsize
    \small #1. #2\par
  \else
    \global \@minipagefalse
    \hb@xt@\hsize{\hfil\box\@tempboxa\hfil}%
  \fi
  \vskip\belowcaptionskip}

\renewenvironment{thebibliography}[1]{%
\begin{oldthebibliography}{#1}%
\small%
\raggedright%
\setlength{\itemsep}{5pt plus 0.2ex minus 0.05ex}%
}%
{%
\end{oldthebibliography}%
}

\begin{document}


\title{\boldmath An Unsupervised Deep-Learning Method 
for Fingerprint Classification: 
the CCAE Network and the Hybrid Clustering Strategy}

\author[a]{Yue-Jie Hou,}
\author[a]{Zai-Xin Xie,}
\author[a,1]{Jian-Hu,}\note{hujian@dali.edu.cn. Corresponding author.}
\author[b,2]{Yao-Shen,}\note{shenyaophysics@hotmail.com. Corresponding author.} 
\author[a,3]{and Chi-Chun Zhou}\note{zhouchichun@dali.edu.cn}
\

\affiliation[a]{School of Engineering, Dali University, Dali, Yunnan 671003, P.R. China}
\affiliation[b]{School of Criminal Investigation, People's Public Security University of China, Beijing 100038, P.R. China}










\abstract{The fingerprint classification is an important and effective method 
to quicken the process and improve the accuracy in 
the fingerprint matching process. 
Conventional supervised methods need a large amount 
of pre-labeled data and thus consume immense human resources. 
In this paper, we propose a new and efficient unsupervised deep 
learning method that can extract fingerprint features and classify 
fingerprint patterns 
automatically.
In this approach, a new model named constraint convolutional auto-encoder 
(CCAE) is used to extract fingerprint 
features and a hybrid clustering strategy is applied to 
obtain the final clusters. 
A set of experiments in the NIST-DB4 dataset shows that the 
proposed unsupervised method exhibits the efficient performance 
on fingerprint classification.
For example, the $CCAE$ achieves an accuracy of $97.3\%$ on only $1000$ 
unlabeled fingerprints in the NIST-DB4. 
The code is available at 
\href{https://github.com/HouYueJie/The-CCAE-Network-and-the-Hybrid-clustering-method}{Github}.
}

\maketitle
\flushbottom


\section{Introduction}
Human fingerprint carries biometric information and is the most 
significant and traditional biometric identification technology 
\cite{cummins1940purkinje,hong1998integrating,saferstein2007criminalistics}.
Due to fingerprints' uniqueness and 
invariance \cite{osterburg1977development,Dass2009Individuality}, 
human fingerprint identification or matching technique are applied to 
the matching of the prints 
not only in criminal evidences \cite{block1931fingerprint,nath1991fingerprint,cole2009suspect} 
but also in authentication, customs transit, 
and public transportation systems 
\cite{han2004study,Ross2006Handbook,watanabe2007user,win2020fingerprint}. 
Thus, fingerprint matching or fingerprint identification technology develops rapidly \cite{maltoni2009handbook,jain2010fingerprint}. 

However, the fingerprint data is very large and messy \cite{jiang2006fingerprint}. 
For example, one has to make comparison between 
the suspect's fingerprint and the huge amount of candidates' 
fingerprints to collect criminal 
evidences \cite{block1931fingerprint,nath1991fingerprint,cole2009suspect}.
According to different conditions of the crime scene, fingerprints are 
usually incomplete and deformed \cite{ross2004biometric}. 
This lead to tremendous difficulties in the fingerprint matching process.
The fingerprint classification is an important and effective method 
to shorten the time and improve the accuracy in 
the fingerprint matching process \cite{kawagoe1984fingerprint,isenor1986fingerprint,galar2015survey}.  
For example, the pre-classification of fingerprint patterns can reduce 
the number of candidate fingerprints 
in the matching stage \cite{Komarinski2006Automated,Wang2016A}.

Fingerprint patterns are basically 
divided into three categories and they are 
arch prints, loop prints, and whorl prints \cite{henry1913classification}. 
Researchers pay attention to the algorithms of fingerprint classification since 
last century \cite{rao1980type,jaycox1931classification,kamijo1993classifying}. 
Certain algorithms were developed to process fingerprint images
\cite{jain2006pores,wang2010global,paulino2012latent}. 
For example,  in order to distinguish the patterns of fingerprints better, 
features such as the core point, the bifurcation, and 
the ridge ending are defined \cite{Bansal2011Minutiae,Wang2011global}. 
Frequency analysis based on transforms such as Fourier transform 
is used to process fingerprint images 
\cite{jain1999multichannel,karungaru2008classification}.
The graph theory methods, such as the orientation flow 
\cite{cappelli2004state,cao2015latent,ozbayouglu2019unsupervised} and 
the singular point \cite{klimanee2004classification,liu2008fingerprint},
are also introduced.

Recently, the neural network methods are applied in 
the fingerprint classification as an universal algorithm 
\cite{peralta2018use,Minaee2019Fingernet,jian2020lightweight}.
Conventional supervised neural network methods report 
high matching accuracy \cite{karungaru2008classification,jeon2017fingerprint,
peralta2018use,tang2019fclassnet,wu2019fingerprint}. 
For example, Jeon et al. \cite{jeon2017fingerprint} 
reported an accuracy of 98.3$\%$ using VGGNet \cite{simonyan2014very}.
CaffeNet (a variant model of AlexNet \cite{krizhevsky2012imagenet})
reported by Peralta et al. \cite{peralta2018use} has an accuracy of 90.7$\%$. 
Wu et al. \cite{wu2019fingerprint} designed FCTP-Net
whose accuracy was 92.9$\%$. 
Moreover, the neural network 
structures with domain knowledge \cite{tang2019fclassnet} were designed. 
Meanwhile, interpretable neural networks were preferred 
\cite{tang2017fingernet,takahashi2020fingerprint}. 
For example, Yao TANG et al. \cite{tang2019fclassnet} 
proposed a model with domain knowledge including orientation-field and singular-point. 
Furthermore, we collect the performance of existing 
supervised learning fingerprint classification models, 
as shown in Table. \ref{exist_slm}. 
Form Table. \ref{exist_slm}, 
The fingerprint classification task based on supervised learning has achieved high accuracy. 
We evaluate our model of unsupervised learning against the model with the highest accuracy. 

\begin{table}[H]
\centering
\caption{A table of the existing supervised learning methods for fingerprint classification}
\resizebox{\textwidth}{18mm}{
\begin{tabular}{ccccc}
\hline
Works & Methods & Testing sets (size) & Tag number & Accuracy($\%$)\\ \hline
Sen Wang et al. \cite{wang2002fingerprint}   & Directional field + kmeans    & NIST 14 dataset (1000)   & 4       & 89.5\\
Ruxin et al. \cite{Wang2016A}                & SAE + softmax regression   & NIST-DB4 (2000)            & 4       & 91.4\\
Wang-Su et al. \cite{jeon2017fingerprint}    & VGGNet    & FVC 2000, 2002, FVC2004 (100)                & 5       & 98.3\\
Daniel et al. \cite{peralta2018use}          & Variant CaffeNet    & NIST-DB4 (1650)                    & 5       & 90.7\\ 
Fan et al. \cite{wu2019fingerprint}          & FCTP-Net    & NIST DB4, DB9, DB10 (2000)                 & 4       & 92.9\\
Wen et al. \cite{jian2020lightweight}        & Singularity ROI + CNN    & NIST-DB4 (4000)               & 5       & 93.0\\
\hline
\label{exist_slm}
\end{tabular}
}
\end{table}

The key to construct an efficient fingerprint classification algorithm is extracting useful features 
that are invariant and carry the information 
of fingerprint patterns \cite{Bansal2011Minutiae}. 
To capture the features of fingerprint patterns manually 
is difficult \cite{peralta2018use,wu2019fingerprint}.
Conventional supervised neural network methods need a large amount 
of pre-labeled data and thus consume immense human resources. 

In this paper, beyond the supervised method, 
we propose an unsupervised classification approach of deep learning. 
The method can extract fingerprint features and classify 
fingerprint patterns automatically.
In this approach, a new convolutional auto-encoder structure named 
constraint convolutional auto-encoder (CCAE) is used to extract fingerprint 
features. At the same time, a hybrid clustering strategy is used to 
obtain the final clusters. 
A set of experiments in NIST-DB4 dataset shows that,
the proposed unsupervised method exhibits the 
efficient performance on fingerprint classification.
It shows that the proposed 
method, compared with the supervised method, not only need no 
pre-labeled fingerprint images but also performs better.

This paper is organized as follows. 
In Sec. 2, 
we introduce the main method, including  
the data preprocessing, 
the CCAE neural network model structure, and the hybrid clustering strategy respectively. 
In Sec. \ref{Result}, we give the result of the main method using
on the NIST-DB4 fingerprint database. Details such as 
the setting of the experimental environment, 
the specific configuration of the experimental model, and 
the performance evaluation of each models are given.
Conclusions and discussions are given in Sec. \ref{CD}.
Another details are given in the appendix.

\section{Database and the main method}
In this section, we introduce the database and the main method.

\subsection{Database and preprocessing}\label{D_P}
The NIST special database 4 (NIST-DB4) \cite{watson1992nist} is a pre-labeled fingerprint 
data set. The NIST-DB4 contains $4000$ 8-bit gray scale 
fingerprint images ($2000$ pairs) stored in PNG format. 
The image size is $512\times512$ PPI. In this work, instead of using the whole 
images from the NIST-DB4, we only uses $1000$ fingerprint images that are
randomly selected from the NIST-DB4. 

\begin{figure}[H]
\centering
\includegraphics[width=0.50\textwidth]{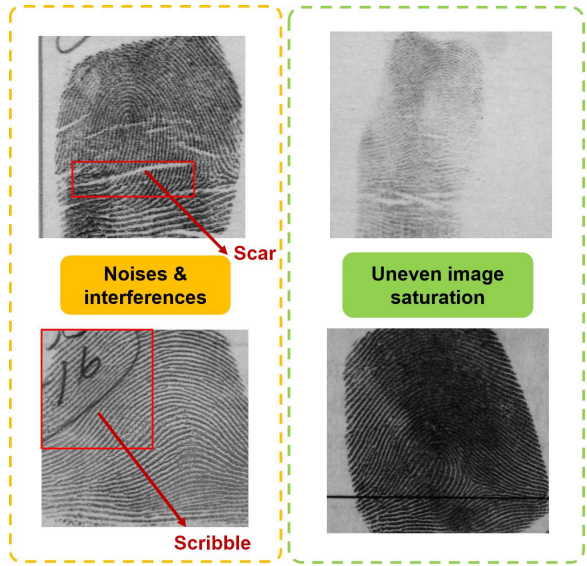}  
	\caption{Examples of images in NIST-DB4 that are affected by noises, interference, and uneven image saturation.}
	\label{image problem}
\end{figure}

Some images in NIST-DB4 are affected by noises, interference, and uneven image 
contrast, as shown in Fig. \ref{image problem}. 
In this section, we apply successive operations on 
the raw fingerprint images in order to obtain images with higher quality.
The preprocessing includes cropped, 
denoising, and binarization of the core area of the fingerprint image.

\begin{figure}[H]
\centering
\includegraphics[width=0.85\textwidth]{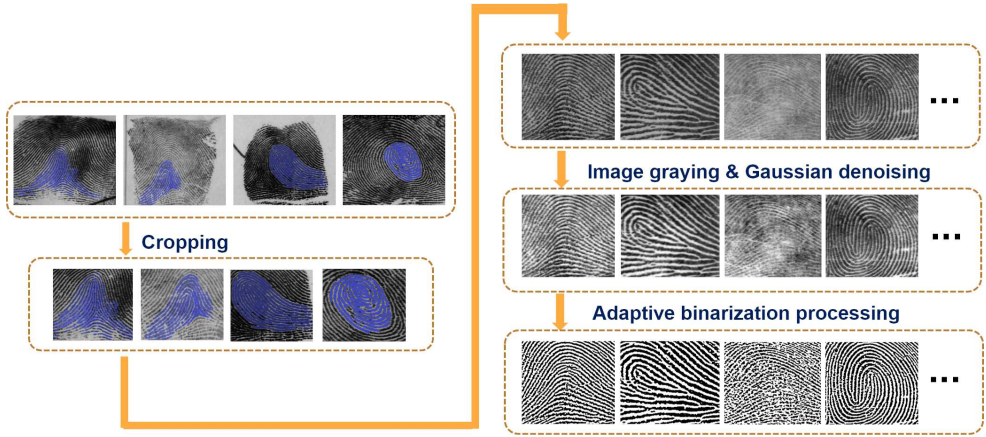}
	\caption{The pre-process: cropping, denoising, and adaptive binarization. The blue area is the central point of the fingerprint}
	:\label{crop}
\end{figure}

\textit{Cropping the core area of the fingerprint image}.
We manually crop out the core area of 
the fingerprint images. 
The cropped process will be replaced by neural network method in the following research.
The original image is cropping from $512\times512$ PPI to $200\times200$ PPI, as shown in Fig. \ref{crop}. 
The cropped image will be resized to $256\times256$ PPI before sending into the $CCAE$ network.

\textit{Denoising and adaptive binarization}. The Gaussian denoising and the 
contrast equalization are applied to remove high 
frequency noise and adjust the image contrast automatically. 
Then, adaptive binarization processing is used to obtain images with higher 
quality, as shown in Fig. \ref{crop}. 

\subsection{The model}\label{the_model}
In this section, we introduce a new convolutional auto-encoder structure named
the constraint convolutional auto-encoder (CCAE).
The $CCAE$ is a variant of the auto-encoder (AE) model. 
A brief review of the AE, including the convolutional auto-encoder (CAE), 
and the variational auto-encoder (VAE),
is given in the appendix. In constructing the loss function,
we use two methods to build the loss function of the $CCAE$. 
The $CCAE$ with these two loss functions are named $CCAE_{a}$ and $CCAE_{b}$
respectively. The $CCAE$, both $CCAE_{a}$ and $CCAE_{b}$, has been proved to be an useful tool
to extract fingerprint features. 
For the sake of convenience, the $CCAE$ represent the $CCAE_{a}$ and the $CCAE_{b}$ models.

\subsubsection{The structure}
\begin{figure}
\centering
\vspace{-1.5em}
\includegraphics[width=0.8\textwidth]{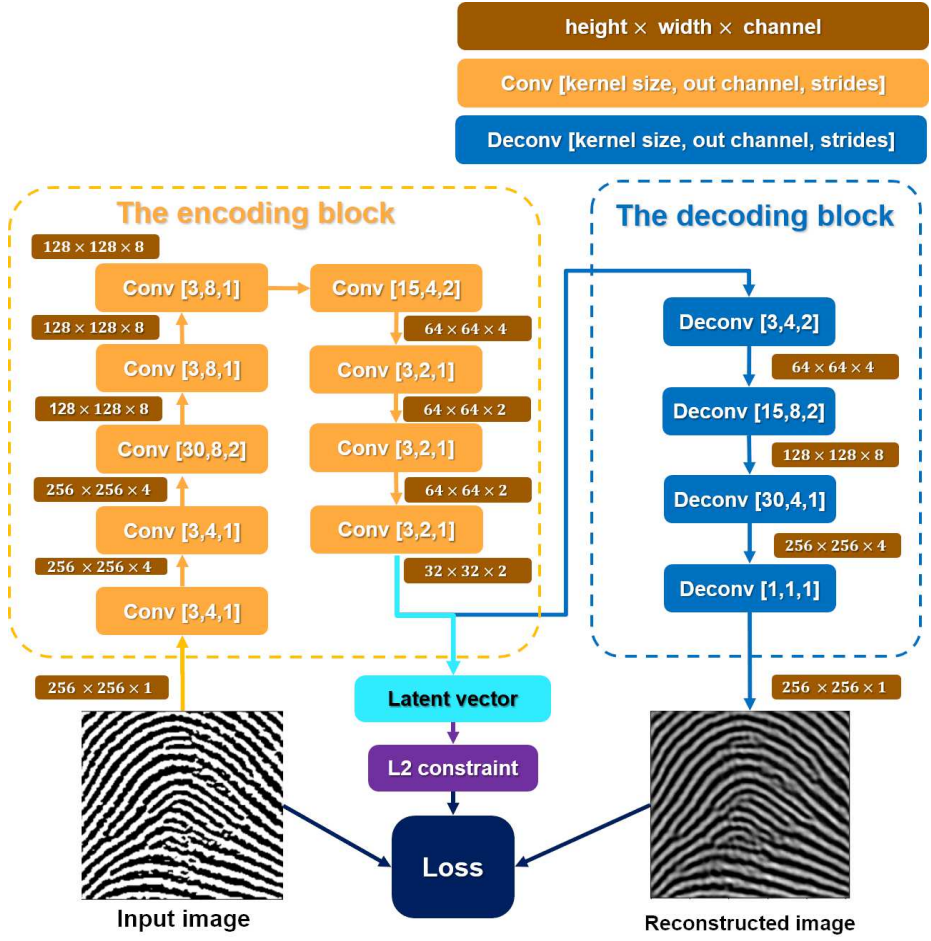}
	\caption{The structure of the $CCAE$.}
	\label{model}
\end{figure}

Instead of using pooling and upsampling layers, the 
$CCAE$ uses convolutional and deconvolution layers only.
Moreover, the convolution layer with big kernel-sizes, 
such as $15\times15$ and $30\times30$, is applied. 
The structure of the $CCAE$ and the parameter setting are 
shown in Fig. \ref{model}. 

\subsubsection{The loss function: the constraint on the latent vector }
To extract effective features in the fingerprint images, 
the model need indeed learn some of the morphological features of the fingerprint, 
but need not learn all the morphological features. 
Because the entire morphological features of the fingerprint
are redundant for the fingerprint classification, and lead to low classification accuracy.
That is, the model should "learn" and at the same time "forget" the features of the fingerprint.
In this section, we introduce two methods to build the loss function of the $CCAE$.  

\textit{$CCAE_{a}$.} In the $CCAE_{a}$, the latent vector parameters $y$ are sampled from a Gaussian 
distribution with mean-value of $0$. 
The constraint is transformed as an additional term in the loss function, as shown in Fig. \ref{AE_CAE_VAE_CCAE}. 
That is
\begin{equation} 
loss=\frac{1}{M}\sum_{i=1}^M\left[\left(x_{i}-x_{i}^{'}\right)^{2}+\frac{1}{N}\sum_{j=1}^N y_{ij}^{2}\right]. \tag{4} \label{con:4}
\end{equation}
where $M$ is the number of fingerprint image, $N$ is the length of latent vector, 
$x_{i}^{'}$ is the decoding data.

\textit{$CCAE_{b}$.} The $CCAE_{b}$ model inspired from the work of Aytekin et al.\cite{aytekin2018clustering}, 
where the loss function reads
\begin{equation}
loss=\frac{1}{M}\sum_{i=1}^M\left(x_{i}-x_{i}^{'}\right)^{2}, 
\quad x_{i}^{'}=D\left({\frac{y_{j}}{\frac{1}{N}\sum_{j=1}^N y_{j}^{2}}}\right) \tag{5} \label{con:5}
\end{equation}
where $D$ is the decoding block proceeding. 
That is the $CCAE$ with Aytekin's loss function, Eq. (\ref{con:5}), is the $CCAE_{b}$ model.

\subsection{The hybrid clustering algorithm}\label{hc}
In this section, we introduce a new clustering strategy called the hybrid clustering algorithm
into the fingerprint clustering. The strategy is first proposed in the previous 
work "Automatic morphological classification of galaxies: convolutional 
auto-encoder and bagging based multi-clustering model" that is under publishing.
The previous work shows that, a single clustering method groups the subsample from a 
single perspective of view and thus will lead to miss-classification. 
By considering 
the clustering result of various kinds of models, 
one can group subsamples from a 
comprehensive view, that is a multi-perspective of view, 
and thus the clustering result is
more reliable. 

Here, the encoding data is used to divide the fingerprint data into 
four categories (arch, whorl, right loop and left loop) by different clustering methods. 
Different clustering methods make different clustering effects. 
The hybrid clustering algorithm, bagging algorithm based on multi-clustering model,
is used to get the optimal clustering results. It shows that the hybrid clustering 
algorithm can cluster the patterns of fingerprint better than 
the traditional single clustering algorithm.

\begin{figure}[H]
\centering
\includegraphics[width=1\textwidth]{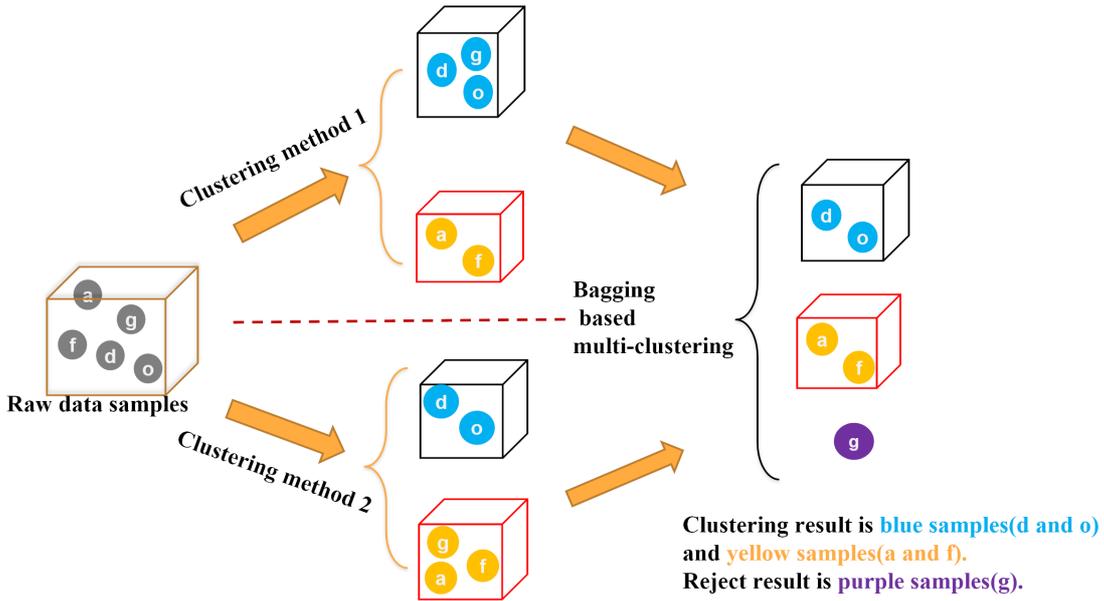}
	\caption{An illustration of the hybrid clustering strategy. 
	For example, the purple ball (g) ball is classified into different clusters, 
	so it is not well clustered and thus rejected. Blue balls, such as balls d and o, 
	and yellow balls, such as balls a and f, are well clustered by both clustering methods.}
	\label{bagging}
\end{figure}

The Hybrid clustering algorithm can improve the accuracy of clustering greatly. 
The algorithm flow chart is shown in Fig. \ref{bagging}. 
Here we use three traditional clustering methods, including 
the Euclidean distance k-mean algorithm \cite{macqueen1967some}, 
the agglomerative (AGG) algorithm \cite{murtagh1983survey}, 
and the balanced iterative reducing and clustering using hierarchies (BIRCH) algorithm \cite{zhang1996birch}.
A brief review of the k-mean, the AGG, and the BIRCH algorithms is given in the appendix.

\begin{figure}[H]
\centering
\includegraphics[width=0.9\textwidth]{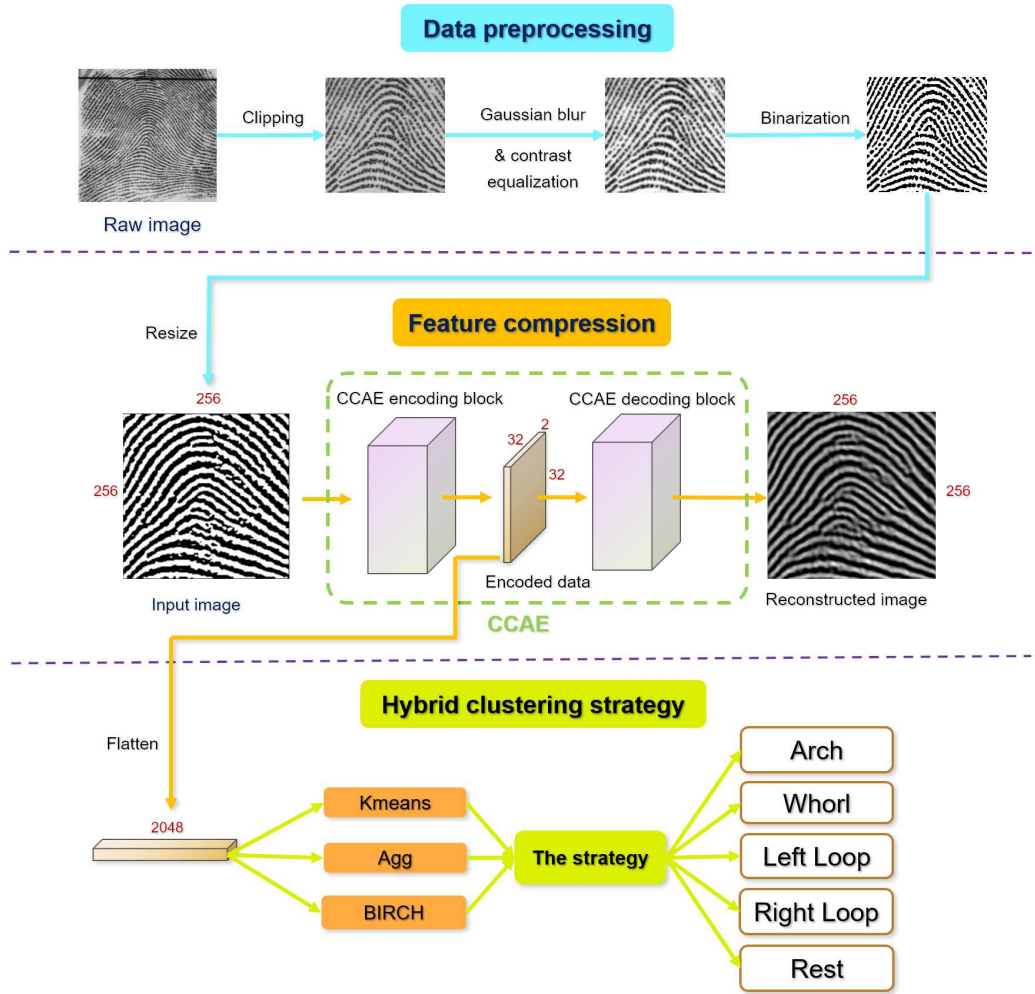}
	\caption{The flow chart.}
	\label{liucheng}
\end{figure}

Although the bagging algorithm based multi-clustering model will result in more rejective data, 
the accuracy will be improved obviously, which will be shown in the following sections. 
For the sake of clarity, Fig. \ref{liucheng} gives an overview of the main method.

\section{Results}\label{Result}
In this section, we provide model evaluation indicators for several models. 
The highlight here is to transfer the evaluation method 
with supervised learning to our unsupervised task. 
All of our models used an Adam \cite{kingma2014adam} optimizer 
with a learning rate of $3\times10^{-4}$ and training 1000 epochs on datasets.

\subsection{Environment settings.} 
The experiment system environment used for learning and testing is set as follows. 
The operating system is Linux Ubuntu 18.04.5. 
The hardware consists of an Intel Xeon CPU E5-2690v3 2.60GHz, 62 GB memory, 
one NVIDIA GeForce GTX 1070Ti GPU, and the Tensorflow 2.4.0 deep learning framework.

\subsection{A comparison with the existing supervised methods}
In order to show the difference between 
the existing supervised methods and our model,
a table is given to show the performance of various approaches. 
See Table. \ref{cf} below.

\begin{table}[H]
\centering
\vspace{-1em}
\caption{A comparison between the existing supervised methods and the $CCAE$. 
our model uses only 1000 randomly selected subsamples from the NIST-DB4 dataset. 
$A_{\; all}$ stands for the all fingerprint type accuracy. 
R.r is the Reject rate, 
which is the percentage of the total number of rejections 
after the hybrid clustering mechanism.}
\begin{tabular}{ccccc}
\hline
\hline
Method  & Training sets (size)  & Testing sets (size)  &$A_{\; all}$($\%$)      &R.r($\%$) \\ \hline
\hline
Ruxin et al. \cite{Wang2016A}       &NIST-DB4 (2000)   & NIST-DB4 (2000)        &91.4          &$\times$     \\
\hline
Fan et al. \cite{wu2019fingerprint} &\begin{tabular}[c]{@{}c@{}}NIST DB4, DB9,\\  DB10 (100000) \end{tabular}   & \begin{tabular}[c]{@{}c@{}}NIST DB4, DB9,\\  DB10 (2000) \end{tabular}        &92.9         &$\times$       \\
\hline
\begin{tabular}[c]{@{}c@{}}The $CCAE_{a}$ and \\ the hybrid clustering\end{tabular}   &None & NIST-DB4 (1000)  &96.4          &5.1            \\
\hline
\begin{tabular}[c]{@{}c@{}}The $CCAE_{b}$ and \\ the hybrid clustering\end{tabular}   &None & NIST-DB4 (1000)  &97.3          &6.3            \\
\hline
\label{cf}
\end{tabular}
\end{table} 
\vspace{-1em} 

Tables \ref{exist_slm} and \ref{cf} shows that
the $CCAE_{a}$ and the $CCAE_{b}$ achieve higher accuracy
than the supervised method without any pre-labeled training dataset.
The number of rejected subsamples in the $CCAE_{a}$ and the $CCAE_{b}$ are only $51$ and $63$
respectively. In other word, the actually size of testing set is 949 and 937 respectively. 

\subsection{The effectiveness of the hybrid clustering algorithm} 
In this section,  in order to show the effectiveness of 
the hybrid clustering algorithm,
we make comparison between the result of 
the single clustering model and the hybrid clustering model, 
as shown in the Table. \ref{v_P} below. 

\begin{table}[H]
\centering
\vspace{-1em}
\caption{Comparison between the hybrid clustering model and 
the single clustering model in terms of their performance. }
\begin{tabular}{ccccc}
\toprule
\multirow {2}{*}{Model} & \multicolumn{2}{c}{Performance evaluation}   \\
\cline{2-3}  
            & $A_{\; all}$($\%$) & R.r($\%$) \\
  \midrule
  The $CCAE_{a-k-means}$                       & 92.6    &  $\times$ \\
  The $CCAE_{a-BIRCH}$                       & 94.9    &  $\times$ \\
  The $CCAE_{a-AGG}$                       & 94.9    &  $\times$ \\
  The $CCAE_{a}$                       & 96.4    &  5.1\\
  The $CCAE_{b}$                       & 97.3    &  6.3\\
  \bottomrule
  \end{tabular}
  \label{v_P}
\end{table}
\vspace{-1em}

Table. \ref{v_P} shows that 
the accuracy of the hybrid clustering model is at least $1.5\%$ higher than 
the single clustering model at the cost of only $6.3\%$ rejected subsamples at most. 
The high-purity clustering result can be used as pre-labeled
training dataset for downstream tasks.

\subsection{The analysis of the $CCAE$}
In this section, 
we further analyze the effectiveness of the $CCAE$. 
Particularly, we discuss the center-crop operation, 
the big convolutional kernel, and L2 constraint on the loss function.

\subsubsection{The center-crop operation} 
The strong prior operation is favorable for model learning features.
We used t-SNE \cite{van2008visualizing}visualization method to directly show 
the improvement brought by strong prior operation.
Fig. \ref{k_c} below shows the results based on the t-SNE visualization.

\begin{figure}[H]
\centering
\includegraphics[width=0.9\textwidth]{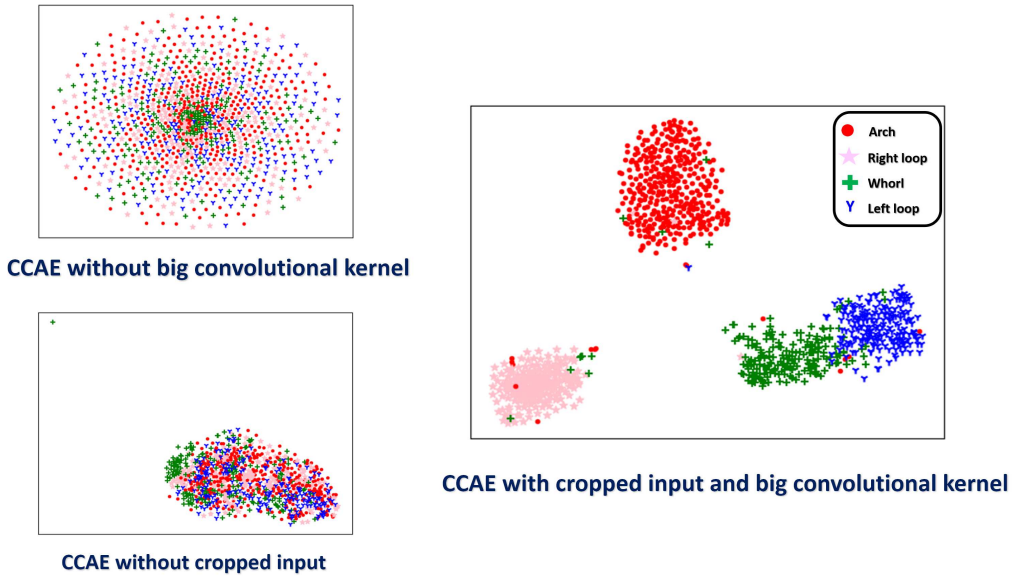}
	\caption{The effect of prior cropped operation on data compression.}
	\label{k_c}
\end{figure}

\subsubsection{The big convolutional kernel} 
We use big convolution kernels for convolution operation, 
whose sizes are 30$\times$30, 15$\times$15 and 9$\times$9. 
The small kernel size is 3$\times$3. 
Fig. \ref{k_c} shows that, 
Base on the $CCAE$, 
the effect of big convolution kernel is critical. Beside, a comparison 
between the $CCAE$ and Aytekin's model is given in the appendix.

\subsubsection{The constraint on the latent vector} 
A constraint on the latent vector is a method that 
designed to "forget" the features 
of the fingerprint. It shows that the $CCAE$ is good at learning the 
required morphological features of the fingerprint.

\begin{table}[H]
\centering
\caption{Comparison of different model in terms of their performance.
$A_{\; all}$ stands for the accuracy, $P_{\; average}$ stands for the average precision,
and $R_{\; average}$ stands for the average recall. The detail of the calculation of $A_{\; all}$ and 
e.t.c., is given in the appendix. 
The cropped CAE model is the big convolution kernel CAE model with the cropped input image.}
\begin{tabular}{ccccccc}
\toprule
\multirow {2}{*}{Model} & \multicolumn{4}{c}{Performance evaluation}   \\
\cline{2-6}  
            & $A_{\; all}$($\%$)      & $P_{\; average}$($\%$)  &  $R_{\; average}$($\%$) & $F1_{\; average}$  & R.r($\%$) \\
  \midrule
  The cropped CAE                 & 95.5       & 94.3     & 95.2    & 0.94  &5.0\\
  $CCAE_{a}$                      & 96.4       & 95.8     & 96.4    & 0.96  &5.1\\
  $CCAE_{b}$                      & 97.3       & 97.0     & 97.3    & 0.97   &6.3\\
  \bottomrule
  \end{tabular}
  \label{P_e}
\end{table}

It shows that adding Gaussian constraint to the encoding data can improve the performance of clustering 
with an increase of $1\%\sim2\%$. Moreover, 
the $CCAE_{b}$ model clustering result can be further improved by $1\%\sim2\%$.
From Table. \ref{P_e}, even if the $CCAE_{b}$ model is highly accurate, 
the rejection rate is also high. 
This indicates that different l2 constraint methods are suitable for different scenarios, 
including high precision rate scenarios or high recall rate scenarios.

\section{Conclusions and Discussions}\label{CD}
In this paper, a fingerprint classification algorithm which 
use the unsupervised $CCAE$ to learn the fingerprint features
and use the hybrid clustering algorithm to 
clustering fingerprints by their patterns is presented. 
To integrally extract fingerprint structure information, 
we use generative model, rather than discriminative model. 
The purpose of the discriminant model is to 
find the high-dimensional representation of 
the data in the high-dimensional space, 
and then map it to a low-dimensional space for classification and discrimination. 
It may contain incomplete structural characteristics of the data. 
However, the generation model can compress the dimension of the data 
while the generation module preserves the structural information of the data. 
This is a more reasonable way of data compression, 
and can provide higher robustness and interpretability for 
downstream data matching and data classification.

We have following main contributions in this paper. 
Instead of using the popular supervised deep learning method, 
(1) we propose a novel unsupervised model, the $CCAE$, to learn the fingerprint features automatically
and (2) apply the hybrid clustering strategy to obtain the final groups. 
It shows that the proposed 
method, compared with the supervised method, not only need no 
pre-labeled fingerprint images but also performs better. 

In the future work, different network structure will be considered, 
such as the residual \cite{K.He2016Deep} and the inception \cite{szegedy2015going}. 
Furthermore, we will migrate the model to various domains, 
such as palm prints, faces, celestial bodies, medical imaging, and so on. 

\section{Acknowledgments}
We are deeply indebted to Prof. Wu-Sheng Dai for his enlightenment and encouragement. 
We are very indebted to Prof. Guan-Wen Fang and Yong-Xie for their encouragements. 
This work is supported by Yunnan Youth Basic Research Projects (202001AU070020 and 202001AU070022) 
and Doctoral Programs of Dali University (KYBS201910).

\section{Appendix}
\subsection{A brief review of the AE, the CAE, and the VAE}
\begin{figure}
\centering
\includegraphics[width=1\textwidth]{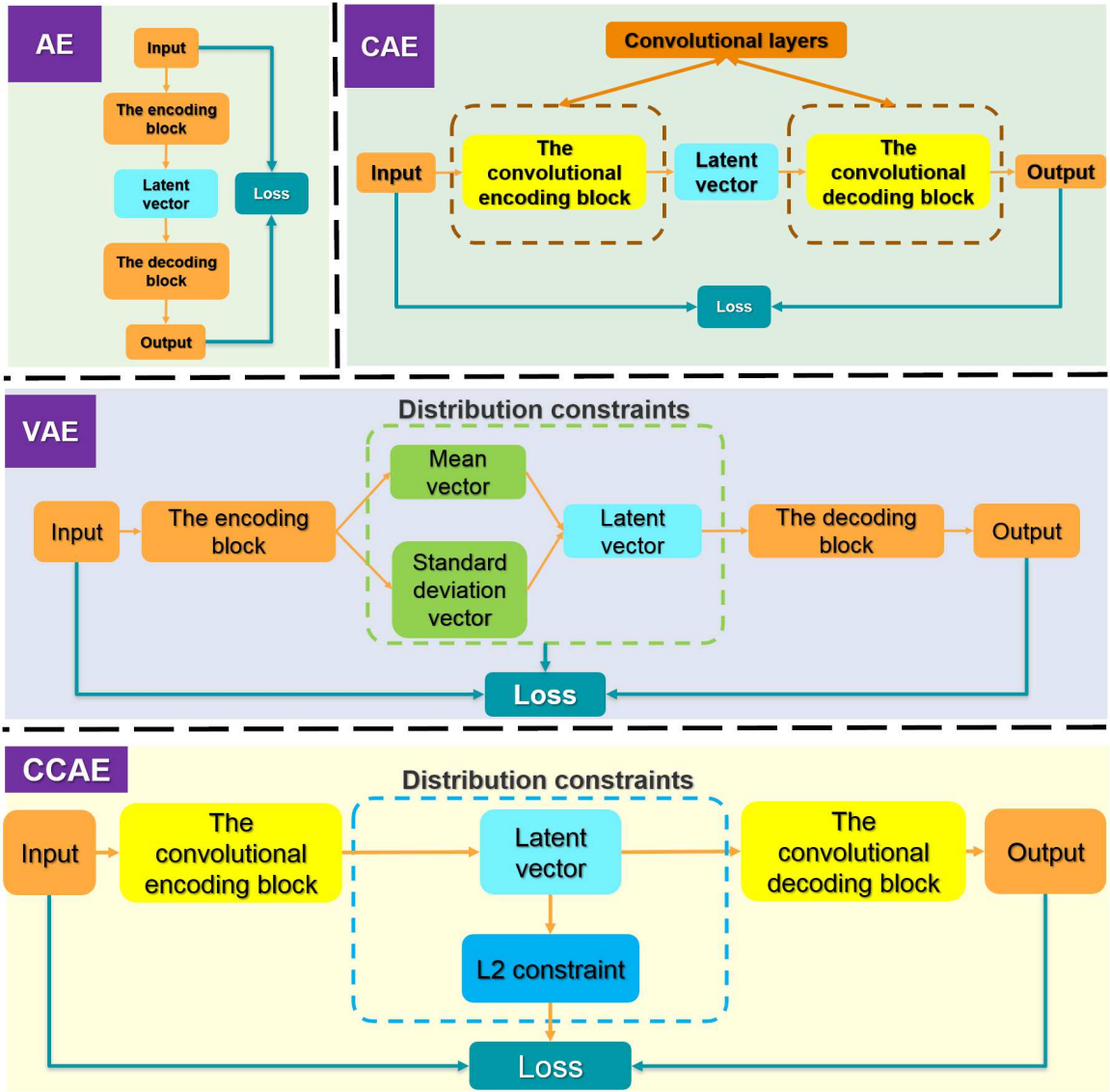}
	\caption{The AE model and the variant AE model}
	\label{AE_CAE_VAE_CCAE}
\end{figure}

The $CCAE$ is a variation of the convolutional auto-encoder (CAE) and the variational auto-encoder (VAE). 
In this section, we give a brief review of the auto-encoder (AE), 
the CAE, and the VAE. 
Their models are shown in Fig. \ref{AE_CAE_VAE_CCAE}.

\textit{AE}. The AE is an 
efficient method to extract important informations from the raw data \cite{Baldi2012autoencoders}. 
By the encoding block and the decoding block, which are composed of multiple fully connected layers, 
the AE reduces the dimension of the input and at the same time preserves the main information. 
Usually, the encoding block can be regarded as  
a nonlinear mapping $f_{encode}\left(x\right)$, that maps the input data
$x$ to the latent vector parameters $y$. For example, 
\begin{equation} 
y = f_{encode}\left(x\right) = \sigma\left(Wx+b\right), \tag{1}  \label{con:1}
\end{equation}
where $\sigma\left(x\right)$ is the nonlinear activation function,  
$W$ and $b$ are parameter and weight of the encoding block. 
The decoding block, another nonlinear mapping,
maps the latent vector parameters $y$ back to the reconstructed data $x^{\prime}$. 
By adjusting the weights both in the encoding and decoding blocks, 
the AE minimizes the differences between the reconstructed data $x^{\prime}$ 
and input data 
$x$.

\textit{CAE}. The convolutional operation is 
an effective method to extract features of
images. For example, the Lenet-5 
network \cite{Lecun1998Gradient} and 
the AlexNet \cite{krizhevsky2012imagenet} are 
the variations of the convolutional neural networks, 
these two networks enhanced the image classification efficiency enormously. 
In a conventional auto-encoder (CAE), 
the encoding block of the AE is 
replaced by a convolutional neural networks, which makes the CAE efficient 
at compressing the dimension of images. 

\textit{VAE}. It has been shown that, applying a constraint on the hidden
parameters $y$ helps the AE to learn more effective features \cite{kingma2013auto}.
For example, in the VAE, the hidden
parameters $y$ are usually sampled from the Gaussian distribution.
Aytekin et al.\cite{aytekin2018clustering} sampled the latent vector 
parameters from the unit ball space.

\textit{Resize}. To compare with CAE model, we set up the resize model, 
which adjusts the original $512\times512\times3$ image to $32\times32\times2$ 
directly with Lanczos interpolation 
algorithm \cite{duchon1979lanczos,fadnavis2014image}.
From the perspective of the clustering results, 
we prove that there is no difference 
between the full convolutional auto-encoder and the pure scaling operation. 
This means that full CAE is not 
enough to achieve the learning feature effect. 

\subsection{A brief review of the conventional clustering algorithm}

\textit{The k-means algorithm: a brief review.} 
The k-means algorithm\cite{macqueen1967some} divides the data set samples into 
k cluster classes by different distance formulas. 
The cluster center is obtained by initializing the mean vector at the beginning. 
Through the greedy strategy, 
the distance between the sample and the cluster center is minimized, 
at the same time, the cluster center is updated. 
Finally, clustering results can be obtained.

\textit{The AGG and the BIRCH algorithms: a brief review.} 
The AGG algorithm \cite{murtagh1983survey} and the BIRCH algorithm \cite{zhang1996birch} 
belong to hierarchical clustering algorithm, 
which divides the data set at different levels into form a tree-like structure. 
The AGG algorithm is a bottom-up aggregation strategy. 
Firstly, each sample in the data set is regarded as an initial cluster, 
and then, in each step of the algorithm running, 
the two clustering clusters closest to each other are found to merge. 
The merging process is repeated until reach the preset number of cluster clusters. 
The BIRCH algorithm uses the clustering feature (CF) tree to perform hierarchical clustering. 
The algorithm builds a CF tree based on input data first. 
Then, the clustering algorithm and the outlier processing on leaf nodes are conducted. 
At the end of clustering, each leaf node becomes a cluster of a sample set.

\subsection{The structure of Aytekin's model}

\begin{figure}[H]
\centering
\includegraphics[width=0.9\textwidth]{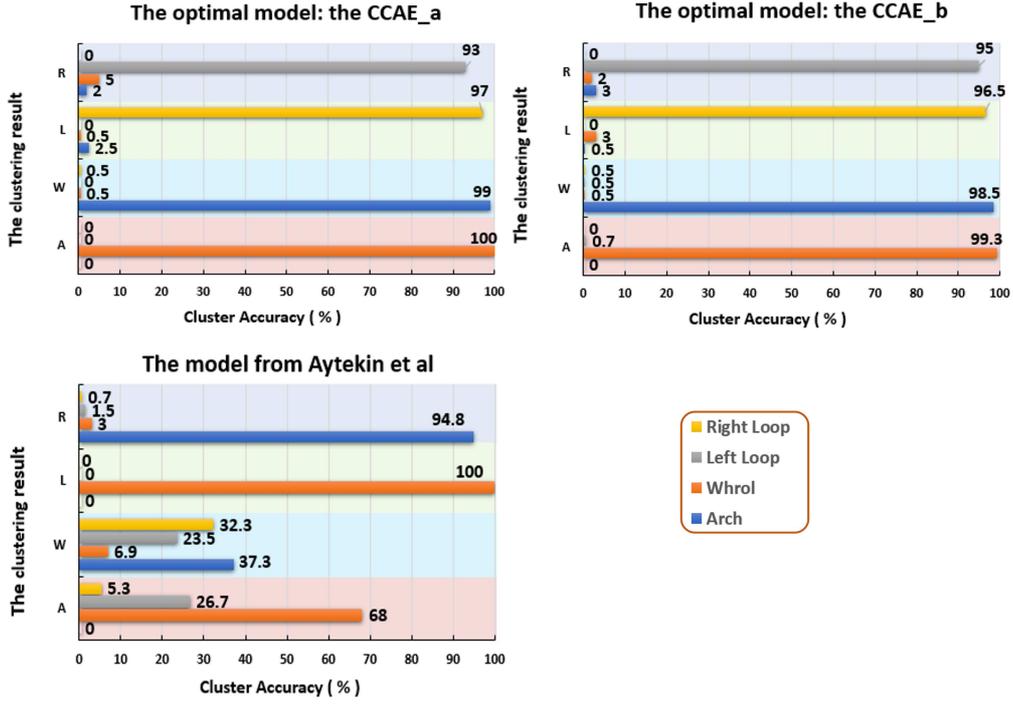}
	\caption{Comparison of clustering effect between the Aytekin's model and the $CCAE$ model.}
	\label{Ay}
\end{figure}

The loss function of the $CCAE_{b}$ model is inspired form the Aytekin's model.
In this section, we make a comparison between the $CCAE$ model and the Aytekin's model.
Fig. \ref{Ay} gives the result and Table. \ref{different_model}
gives the structure of the Aytekin's model

\begin{table}[H]
\small
\centering
\caption{The Aytekin's model and comparison with the $CCAE_{a}$ model and the $CCAE_{b}$ model.}
\begin{tabular}{|c|c|c|}
\hline
\diagbox{Layer}{Method}{Model}  & The Aytekin's model &Output size \\ \hline
layer1       & Conv[5,32,2] &128$\times$128$\times$32    \\ \hline
layer2      & Conv[5,64,2]  &64$\times$64$\times$64  \\ \hline
layer3      & Conv[3,128,2] &32$\times$32$\times$128  \\ \hline
layer4      & Dense[2048]   &2048$\times$1 \\ \hline
layer5       & Deconv[3,128,2] &64$\times$64$\times$128   \\ \hline
layer6       & Deconv[5,64,2]&128$\times$128$\times$64   \\ \hline
layer7       & Deconv[5,32,2]&256$\times$256$\times$32   \\ \hline
layer8       & Deconv[3,1,1]&256$\times$256$\times$1  \\ \hline

\end{tabular}
\label{different_model}
\end{table}

\subsection{The calculation of evaluation index}
We calculate the overall accuracy of the fingerprints, the average precision, 
the average recall, and the average F1 score of the model. 
These parameters respectively represent the overall accuracy of the model, 
the intra-class clustering accuracy, the inter-class clustering accuracy, 
and the intra-class and the inter-class accuracy harmonic average. 
The F1 score is a good indicator of model accuracy. 
The closer its value goes to 1, the better the model. 
Their calculation formula is shown as:
\begin{equation} 
Accuracy_{\; all} = \frac{\sum_{\; class} TP_{\; class}}{\sum_{\; class} Total_{\; class}}, \tag{6} \label{con:6}
\end{equation}
\begin{equation} 
Precision_{\; average} = \frac{1}{N}\sum_{\; class} \frac{TP_{class}}{FP_{\; class}+TP_{\; class}}, \tag{7} \label{con:7}
\end{equation}
\begin{equation} 
Recall_{\; average} = \frac{1}{N}\sum_{\; class}\frac{ TP_{\; class}}{TP_{\; class}+FN_{\; class}}, \tag{8} \label{con:8}
\end{equation}
\begin{equation} 
F1_{\; average} = \frac{1}{N}\sum_{\; class}\frac{2\; \times \;  Precision_{\; class}\;  \times \; Recall_{\; class}}{Precision_{\; class}\;+\;Recall_{\; class}}, \tag{9} \label{con:9}
\end{equation}
where $class$ is the fingerprint pattern, 
it contains arch (arch and tented arch), whorl, left loop, and right loop. 
$TP$ is the correct number of clustering categories and 
$FP$ is the number of the fingerprint intra-class error clustering. 
$FN$ is the number of the fingerprint inter-class error clustering. 
$N$ is the number of the fingerprint patterns. 
The $Precision_{\; class}$ and the $Recall_{\; class}$ are precision and recall within a single class. 
This approach is a multi-classification problem, 
so we use the average values to evaluate performance of the model.












\end{document}